\documentclass[11pt,a4paper]{article}
\usepackage[hyperref]{acl2020}
\usepackage{times}
\usepackage{latexsym}

\usepackage{graphicx}
\usepackage{amsmath}
\usepackage{mathrsfs}
\usepackage[]{xcolor}
\usepackage{etoolbox}
\usepackage{algorithmic}
\usepackage{algorithm}
\usepackage{dirtytalk}

\usepackage{color}
\usepackage{amssymb}
\usepackage{times}
\usepackage{helvet}
\usepackage[normalem]{ulem}
\usepackage{courier}
\usepackage{array,colortbl,multirow,multicol,booktabs,ctable}
\usepackage{graphicx}
\usepackage[inline]{enumitem}
\usepackage{microtype}

\aclfinalcopy % Uncomment this line for the final submission
\def\aclpaperid{2221} %  Enter the acl Paper ID here

\title{Discourse-Aware Neural Extractive Text Summarization
}

\author{Jiacheng Xu\thanks{\,\, Most of this work was done when the first author was
an intern at Microsoft.} \textsuperscript{\rm 1}, Zhe Gan\textsuperscript{\rm 2}, Yu Cheng\textsuperscript{\rm 2}, Jingjing Liu\textsuperscript{\rm 2} \\
  \textsuperscript{\rm 1}The University of Texas at Austin \quad \textsuperscript{\rm 2}Microsoft Dynamics 365 AI Research\\
\texttt{jcxu@cs.utexas.edu; \{zhe.gan,yu.cheng,jingjl\}@microsoft.com } \\}

\date{}

\begin{document}
\maketitle
\begin{abstract}
Recently BERT has been adopted for document encoding in state-of-the-art text summarization models. 
However, sentence-based extractive models often result in redundant or uninformative phrases in the extracted summaries. Also, long-range dependencies throughout a document are not well captured by BERT, which is pre-trained on sentence pairs instead of documents.
To address these issues, we present a discourse-aware neural summarization model - \textsc{DiscoBert}\footnote{Code, illustration and datasets are available at: \href{https://github.com/jiacheng-xu/DiscoBERT}{https://github.com/jiacheng-xu/DiscoBERT}.}.
\textsc{DiscoBert} extracts sub-sentential discourse units (instead of sentences) as candidates for extractive selection on a finer granularity.
To capture the long-range dependencies among discourse units, structural discourse graphs are constructed based on RST trees and coreference mentions, encoded with Graph Convolutional Networks.
Experiments show that the proposed model outperforms state-of-the-art methods by a significant margin on popular summarization benchmarks compared to other BERT-base models.
\end{abstract}
\section{Introduction}

% \yu{We can merge/shorten the first two paragraphs into one? I guess reviewers in this area do not need much basic context}
Neural networks have achieved great success in the task of text summarization  \cite{nenkova2011automatic,yao2017recent}. There are two main lines of research: abstractive and extractive.
While the abstractive paradigm \cite{Rush_A_2015,See_Get_2017,celikyilmaz-etal-2018-deep,sharma2019entity} focuses on generating a summary word-by-word after encoding the full document, the extractive approach \cite{cheng-lapata-2016-neural,Zhou_Neural_2018,Narayan_Ranking_2018} directly selects sentences from the document to assemble into a summary. 
The abstractive approach is more flexible and generally produces less redundant summaries, while the extractive approach enjoys better factuality and efficiency \cite{CaoEtAl2018fact}. 
% Motivation I
% \jj{This paragraph should highlight why sentence-level is not sufficient, to explain the motivation of discourse-unit-level approach.}

Recently, some hybrid methods have been proposed to take advantage of both, by designing a two-stage pipeline to first select and then rewrite (or compress) candidate sentences \cite{Chen_Fast_2018,Gehrmann_Bottom_2018,Zhang_Neural_2018,xu-durrett-compression}.
%  proposed a neural pipeline framework to conduct text compression based on constituency tree pruning after the sentence selection module. 
% Compression or rewriting aims to discard uninformative phrases from the originally selected sentences.
% However, using sentence as the minimal selection unit inevitably limits the model's ability of reducing redundancy, and merely relying on the next compression step for potential remedy may not be optimal. 
Compression or rewriting aims to discard uninformative phrases in the selected sentences. However, most of these hybrid systems suffer from the inevitable disconnection between the two stages in the pipeline.
% , resulting in limited performance improvement.

% and generating informative summaries

%such two-stage system requires the coupling of two functionally-different modules (i.e., sentence extraction and rewriting) and is fragile as the error propagates during inference, thus only marginal improvements are shown when compared with their extractive counterparts.
%
%most state-of-the-art hybrid systems only show marginal improvement over their extractive counterparts, because 

% [width=0.48125\textwidth] for ACL
% [width=0.473\textwidth] for AAAI
\begin{figure}[t]
\centering
\small
% \justify
% {Example: [Police]\textsubscript{1} [investigating the case]\textsubscript{2}  [learned]\textsubscript{3}  [where 36-year-old Vittorio Arrigoni was being held]\textsubscript{4}  [and went to the location,]\textsubscript{5}  [where they found the body,]\textsubscript{6}  [the statement said.]\textsubscript{7} }
\includegraphics[width=0.473\textwidth]{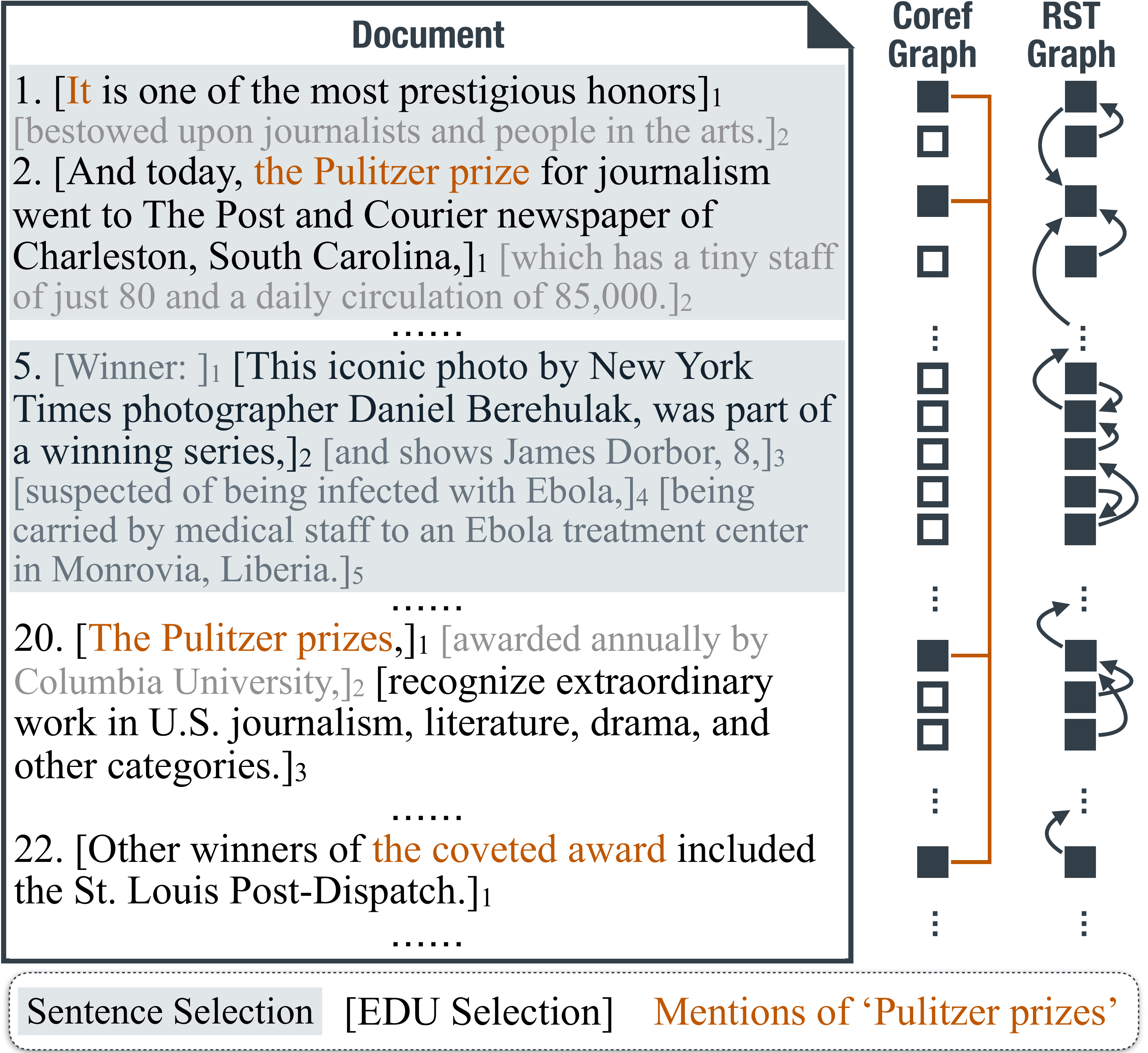}
\caption{Illustration of \textsc{DiscoBert} for text summarization. 
Sentence-based \textsc{Bert} model (baseline) selects whole sentences 1, 2 and 5. 
The proposed discourse-aware model \textsc{DiscoBert} selects EDUs \{1-1, 2-1, 5-2, 20-1, 20-3, 22-1\}.
% , which avoids unnecessary details and generates a more concise summary. 
The right side of the figure illustrates the two discourse graphs we use: ($i$) Coref(erence) Graph (with the mentions of `Pulitzer prizes' highlighted as examples); and ($ii$) RST Graph (induced by RST discourse trees).
}
\label{fig:overview}
%\vspace{-5mm}
\end{figure}
% Police investigating the case learned where 36-year-old Vittorio Arrigoni was being held and went to the location, where they found the body, the statement said.

% \jcxu{TODO better name here?}.

% Motivation II
Meanwhile, modeling long-range context for document summarization remains a challenge \cite{xu-etal-2016-cached}. Pre-trained language models \cite{devlin-etal-2019-bert} are designed mostly for sentences or a short paragraph, thus poor at capturing long-range dependencies throughout a document.
% With the recent success of pre-trained language models (LMs) \cite{devlin-etal-2019-bert}, the encoding of input document has been greatly improved.
% However, since pre-trained LMs are mostly designed for target sentence pairs or short paragraphs, they perform poorly at capturing long-range dependencies among sentences.
%and operated on the word level, 
% how to utilize these LMs to capture long-range dependency between sentences still remains unclear.
Empirical observations \cite{liu-lapata-2019-text} show that adding standard encoders such as LSTM or Transformer \cite{vaswani2017attention} on top of \textsc{Bert} to model inter-sentential relations does not bring in much performance gain.

In this paper, we present \textsc{DiscoBert}, a discourse-aware neural extractive summarization model built upon \textsc{Bert}. 
% \jj{Should explain what compression means.} 
To perform compression with extraction simultaneously and reduce redundancy across sentences, we take Elementary Discourse Unit (EDU), a sub-sentence phrase unit originating from RST \cite{mann1988rhetorical,carlson-etal-2001-building}\footnote{We adopt RST as the discourse framework due to the availability of existing tools, the nature of the RST tree structure for compression, and the observations from \citet{louis-etal-2010-discourse}. Other alternatives includes Graph Bank \cite{wolf2005representing} and PDTB \cite{miltsakaki2004penn}. } as the minimal selection unit (instead of sentence) for extractive summarization. Figure~\ref{fig:overview} shows an example of discourse segmentation, with sentences broken down into EDUs (annotated with brackets). 
% The Sentence Selection is realized by a sentence-based BERT model, and the EDU Selection is achieved by our model. 
By operating on the discourse unit level, our model can discard redundant details in sub-sentences, therefore retaining additional capacity to include more concepts or events, leading to more concise and informative summaries.
% \jj{Refer to Figure 1 for examples.}

%model with EDU for compression.
% \jj{Move to later} \jj{Should we explain what oracles are?} \jcxu{move to later}

% Furthermore, to structually capture document-level long-distance dependency, we propose a graph-based approach to leverage intra-sentence discourse relations among EDUs.
Furthermore, we finetune the representations of discourse units with the injection of prior knowledge to leverage intra-sentence discourse relations.  
More specifically, two discourse-oriented graphs are proposed: RST Graph $\mathcal{G}_{R}$ and Coreference Graph $\mathcal{G}_{C}$. Over these discourse graphs, Graph Convolutional Network (GCN) \cite{kipf2016semi} is imposed to capture long-range interactions among EDUs.
% Two types of discourse graphs are proposed: ($i$) a directed RST-based Graph, and ($ii$) an undirected Coreference Graph. 
\emph{RST Graph} is constructed from RST parse trees over EDUs of the document. 
% Rhetorical relations of EDUs, such as contradiction, elaboration, and attribution, are addressed in the RST Graph. 
On the other hand, \emph{Coreference Graph} connects entities and their coreference clusters/mentions across the document. 
% In Figure~\ref{fig:overview}, we show part of the coreference mention cluster of `Pulitzer prize'. 
The path of coreference navigates the model from the core event to other occurrences of that event, and in parallel explores its interactions with other concepts or events.

%redundancy reduction. 
%\zhe{intra-sentence is not accurate? since now we work on discourse units.} 
%Experiments on popular text summarization datasets show that the constructed graphs, together with the GCN model, 
%could further update the representation of discourse units from BERT, and 
%boost summarization performance. 
%capture longer-range context for document summarization, benefiting from the injected prior linguistic knowledge. 

% Challenge:
% 1) global context is not fully captured
% 2) redundancy compression
% complicated two-stage model for text compression and doesn't work well.

The main contribution is threefold:
($i$) We propose a discourse-aware extractive summarization model, \textsc{DiscoBert}, which operates on a sub-sentential discourse unit level to generate concise and informative summary with low redundancy. 
% treating EDUs (instead of sentences) as the minimal selection unit to provide a fine-grained granularity for extractive selection, while preserving the grammaticality and fluency of generated summaries. 
%To the authors' best knowledge, this is the first fully end-to-end neural summarization model that simultaneously achieves extraction and compression.
%\jj{what is text compression model? Extractive?} 
($ii$) We propose to structurally model inter-sentential context with two types of discourse graph.
% two discourse graphs, and use a graph-based approach to model the inter-sentential context based on discourse relations among EDUs.
%for document-level saliency estimation \jj{What is saliency estimation? Never mentioned in the previous sections}. GCN is further applied to capture long-range dependencies in a document for better summarization.
($iii$) \textsc{DiscoBert} achieves new state of the art on two popular newswire text summarization datasets, outperforming other \textsc{Bert}-base models. 
  % Human evaluation and analysis.
%which ($i$) takes fine-grained discourse unit instead of the sentence as the basis for candidate selection, and ($ii$) uses a graph-based approach to model inter-sentential context in a document based on linguistic relations between discourse units. 

%\input{sec_1disco.tex}
% [width=0.48125\textwidth] for ACL
% [width=0.473\textwidth] for AAAI
\begin{figure}[t]
% \centering
% \justify
% {Example: [Police]\textsubscript{1} [investigating the case]\textsubscript{2}  [learned]\textsubscript{3}  [where 36-year-old Vittorio Arrigoni was being held]\textsubscript{4}  [and went to the location,]\textsubscript{5}  [where they found the body,]\textsubscript{6}  [the statement said.]\textsubscript{7} }
\includegraphics[width=0.48125\textwidth]{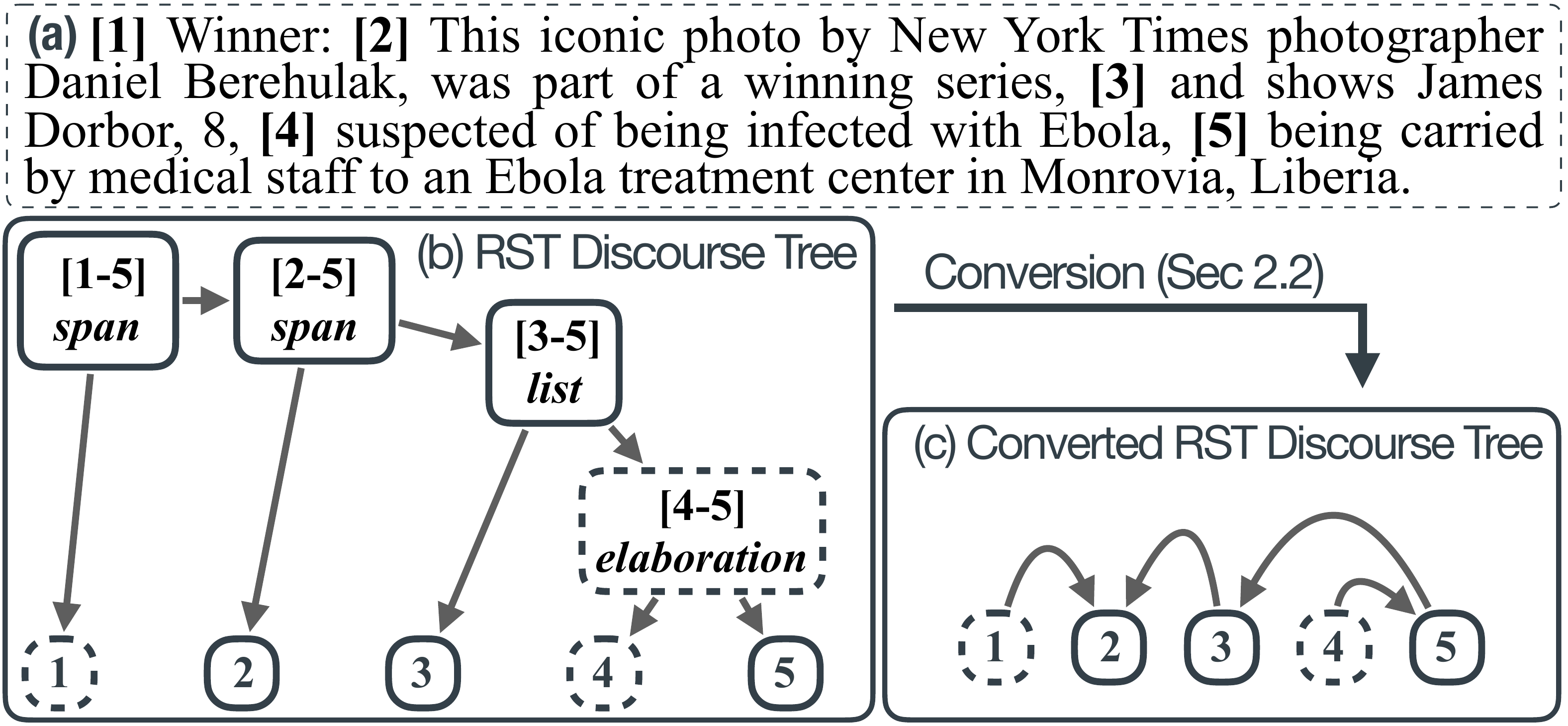}
\caption{Example of discourse segmentation and RST tree conversion.
The original sentence is segmented into 5 EDUs in box (a), and then parsed into an RST discourse tree in box (b). The converted dependency-based RST discourse tree is shown in box (c).
Nucleus nodes including [2], [3] and [5], and Satellite nodes including [2] and [4] are denoted by solid lines and dashed lines, respectively. 
% Leaf nodes with text are EDUs with their property tagged. 
% Non-terminal relation nodes with start and end of the span are shown on the right side \jj{upper or lower?}.
\textit{Relations} are in italic. 
The EDU [2] is the head of the whole tree (span [1-5]), while the EDU [3] is the head of the span [3-5].
% The rightmost \textit{span} node, covering the whole sentence, is the root node of this tree.
}
\label{fig:disco}
%\vspace{-3mm}
\end{figure}
% Police investigating the case learned where 36-year-old Vittorio Arrigoni was being held and went to the location, where they found the body, the statement said.

\section{Discourse Graph Construction}
% \jcxu{goal: two graphs construction. for rst we need to talk about 1) head determination  2) how to convert from RST to dependency (mostly appendix) 3) methods with examples in figure 1. 4) visualization.   for coref graph. mostly current but merge the example }
In this section, we first introduce the Rhetorical Structure Theory (RST) \cite{mann1988rhetorical}, a linguistic theory for discourse analysis, and then explain how we construct discourse graphs used in \textsc{DiscoBert}.
Two types of discourse graph are considered: RST Graph
%\footnote{We will use RST Graph in the rest of paper for simplicity.} 
and Coreference Graph. 
All edges are initialized as disconnected, and connections are later added for a subset of nodes based on RST discourse parse tree or coreference mentions.
% For initialization of both graphs, all edges are disconnected. Connections are then added for a subset of nodes based on RST discourse parse tree or coreference mentions.

\subsection{Discourse Analysis}

%Discourse analysis focuses on inter-sentential relations in a document or conversation. 
% In Rhetorical Structure Theory (RST) \cite{mann1988rhetorical}, the discourse structure of a text can be represented as a tree. 
% The whole document can be segmented into contiguous, adjacent and non-overlapping text spans called Elementary Discourse Units (EDUs). 
%Each EDU is tagged as either \textit{Nucleus} or \textit{Satellite}, which characterizes the document's nuclearity or saliency. \textit{Nucleus} (N) nodes are more central in terms of its content and grammatical/lexical reliance, while \textit{Satellite} (S) nodes are more peripheral.

Discourse analysis focuses on inter-sentential relations in a document or conversation. 
In the RST framework, the discourse structure of text can be represented in a tree format. 
The whole document can be segmented into contiguous, adjacent and non-overlapping text spans called Elementary Discourse Units (EDUs). 
Each EDU is tagged as either Nucleus or Satellite, which characterizes its nuclearity or saliency. 
Nucleus nodes are generally more central, and Satellite nodes are more peripheral and less important in terms of content and grammatical reliance. 
There are dependencies among EDUs that represent their rhetorical relations.

In this work, we treat EDU as the minimal unit for content selection in text summarization. 
Figure~\ref{fig:disco} shows an example of discourse segmentation and the parse tree of a sentence. 
Among these EDUs, rhetorical relations represent the functions of different discourse units. As observed in \citet{louis-etal-2010-discourse}, the RST tree structure already serves as a strong indicator for content selection. On the other hand, the agreement between rhetorical relations tends to be lower and more ambiguous. Thus, we do not encode rhetorical relations explicitly in our model. 
% For example, `elaboration' shows that one EDU describes the detail of the other EDU. `Contrast' means that two EDUs are rhetorically different or opposed to each other.
% `Same\_Unit' reveals the parallel or coexistent relationship between EDUs.
% We show a concrete example in Figure~\ref{fig:disco}, in which the original sentence is segmented into 5 continuous EDUs.

In content selection for text summarization, we expect the model to select the most concise and pivotal concept in the document, with low redundancy.\footnote{For example, in Figure~\ref{fig:disco}, details such as the name of the suspected child in [3], the exact location of the photo in [5], and who was carrying the child in [4], are unlikely to be reflected in the final summary.}
% \jj{what does this mean: the location of the photo, who was...?}\jcxu{this relate to the example in the motivation figure}
However, in traditional extractive summarization methods, the model is required to select a \textit{whole} sentence, even though some parts of the sentence are not necessary.
Our proposed approach can select one or several fine-grained EDUs to render the generated summaries less redundant. This serves as the foundation of our \textsc{DiscoBert} model. 

% So far we treated the EDUs as a new subsentence-level unit for selection. Actually 
\subsection{RST Graph} %$\mathcal{G}_{D}$}
% \jcxu{how do you like it}
% After treating the EDUs as better, fine-grained selection units, we want to further capture the rhetorical structure of the document and build a discourse-aware model for text summarization. 

%As aforementioned, we adopt EDU as the minimal selection unit for our summarization model. Here, we further utilize the discourse trees of a document to capture the rhetorical structure of the document, and build a discourse-aware model for text summarization. 
%In this section, we discuss how the linguistic graphs are constructed. 
% To address the long-range dependency issue in long document encoding, two types of linguistic graphs are constructed. 
% The RST Discourse Graph $\mathcal{G}_{D}$ is a directed graph based on RST, while the Coreference Graph $\mathcal{G}_{C}$ is an undirected graph based on entity coreference clusters and mentions in the document.
%The constructed graphs will be used in the graph encoders of the proposed model. 

% A discourse graph is represented as $\mathcal{G}=(\mathcal{V}, \mathcal{E})$, where $\mathcal{V}$ is the set of all EDUs, and $\mathcal{E}$ is the set of all the connections between EDUs. Let $v_i$ be the $i$-th EDU in the document, and $e_{ij}$ be the edge between the $i$-th and the $j$-th EDUs. 
% Two discourse graphs are constructed: ($i$) RST Graph, and ($ii$) Coreference Graph. 
% For initialization of both graphs, all edges are disconnected. Connections are then added for certain nodes based on RST discourse parse tree or coreference mentions.

%\paragraph{Dependencies among EDUs}
When selecting sentences as candidates for extractive summarization, we assume each sentence is grammatically self-contained. But for EDUs, some restrictions need to be considered to ensure grammaticality. For example, 
Figure~\ref{fig:disco} illustrates an RST discourse parse tree of a sentence, where \say{[2] This iconic ... series} is a grammatical sentence but \say{[3] and shows ... 8} is not. 
We need to understand the dependencies between EDUs to ensure the grammaticality of the selected combinations. The detail of the derivation of the dependencies could be found in Sec~\ref{sec:impl}.

%\paragraph{Construction of Discourse Graph}
The construction of the RST Graph aims to provide not only local paragraph-level but also long-range document-level connections among EDUs. 
We use the converted dependency version of the tree to build the RST Graph $\mathcal{G}_{R}$, by initializing an empty graph and treating every discourse dependency from the $i$-th EDU to the $j$-th EDU as a directed edge, \emph{i.e.}, $\mathcal{G}_{R}[i][j] = 1$.

\subsection{Coreference Graph}  %$\mathcal{G}_{C}$ }
Text summarization, especially news summarization, usually suffers from the well-known `position bias' issue \cite{kedzie2018content}, where most of the key information is described at the very beginning of the document. 
However, there is still a decent amount of information spread in the middle or at the end of the document, which is often ignored by summarization models. 
We observe that around 25\% of oracle sentences appear after the first 10 sentences in the CNNDM dataset.
Besides, in long news articles, there are often multiple core characters and events throughout the whole document. However, existing neural models are poor at modeling such long-range context, especially when there are multiple ambiguous coreferences to resolve.

\begin{algorithm}[t]
\small{
\caption{\small{Construction of the Coreference Graph $\mathcal{G}_{C}$.}}
\label{algo:cm}
\centering
\begin{algorithmic}
\REQUIRE{Coreference clusters $C=\{C_1, C_2, \cdots, C_n\}$; mentions for each cluster $C_i = \{E_{i1}, \cdots, E_{im}\}$.
}
% \STATE \textbf{Input:} Coreference clusters $C=\{C_1, C_2, \cdots, C_n\}$; mentions for each cluster $C_i = \{E_{i1}, \cdots, E_{im}\}$.
\STATE Initialize the Graph $\mathcal{G}_{C}$ without any edge $\mathcal{G}_{C}[*][*]=0$.
\FOR{$i=0$ to $n$}
\STATE Collect the location of all occurences $\{E_{i1}, \cdots, E_{im}\}$ to $L = \{l_1, \cdots, l_m \}$.
\FOR{$j=1$ to $m$, $k=1$ to $m$}
% \FOR{$j=1$ to $m$}
% \STATE  $\mathcal{G}_{C}[j][j] = 1$ 
% \FOR{$k=1$ to $m$}
\STATE  $\mathcal{G}_{C}[j][k] = 1$ 
% \ENDFOR
\ENDFOR
\ENDFOR
\RETURN{Constructed Graph $\mathcal{G}_{C}$. }
\end{algorithmic}}
\end{algorithm}
%\vspace{-2mm}

\begin{figure*}[t!]
\centering
\small
\includegraphics[width=\textwidth]{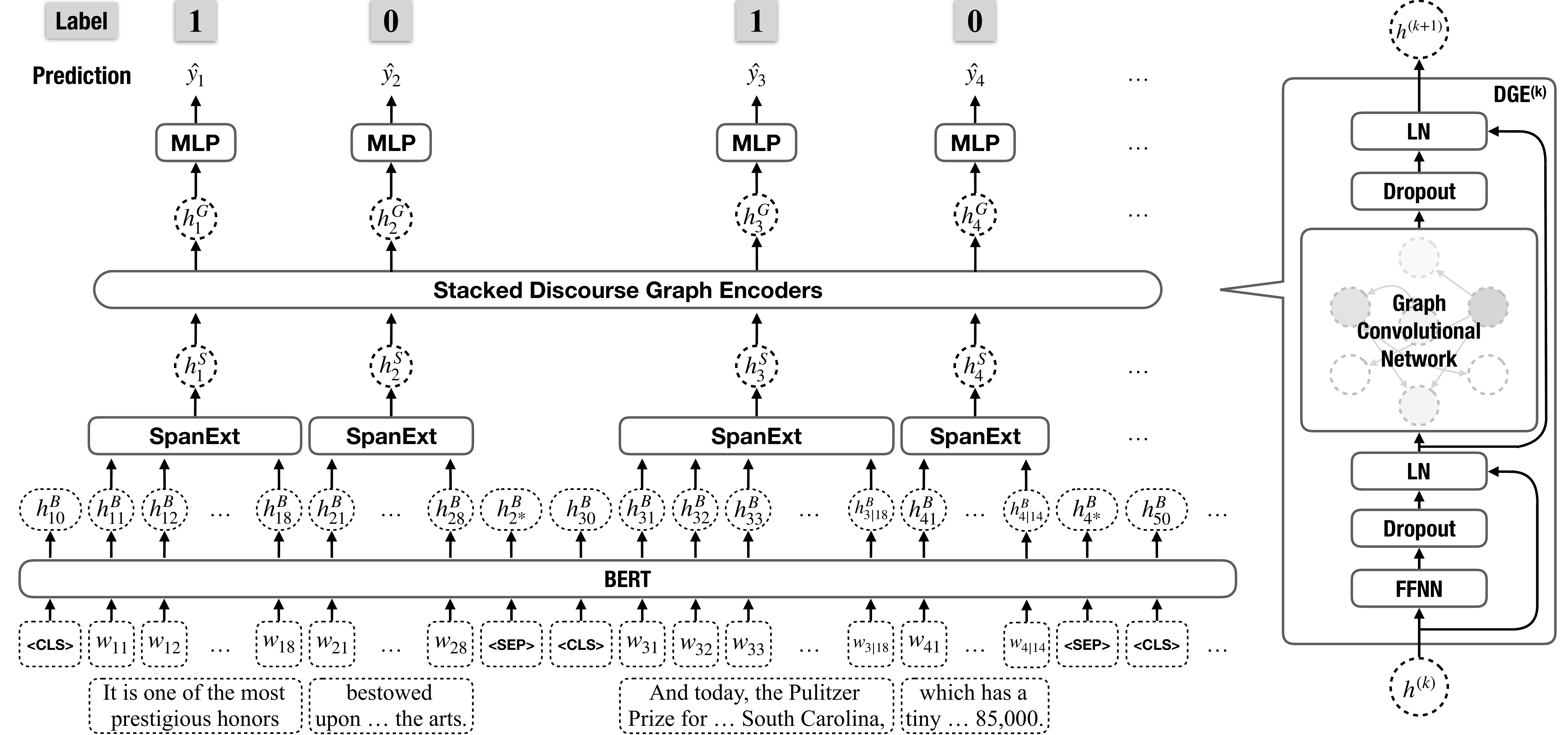}
\caption{(Left) Model architecture of \textsc{DiscoBert}. The Stacked Discourse Graph Encoders contain $k$ stacked DGE blocks. (Right) The architecture of each Discourse Graph Encoder (DGE) block.  }
\label{fg:model}
%\vspace{-3mm}
\end{figure*}

To encourage and guide the model to capture the long-range context in the document, we propose a Coreference Graph built upon discourse units. % as one of our discourse graphs. 
Algorithm~\ref{algo:cm} describes how to construct the Coreference Graph. 
We first use Stanford CoreNLP \cite{Manning_The_2014} to detect all the coreference clusters in an article. For each coreference cluster, all the discourse units containing the mention of the same cluster will be connected. This process is iterated over all the coreference mention clusters to create the final Coreference Graph. 
% \jcxu{TODO: finalize figure 1 and replace the example here.}

Figure~\ref{fig:overview} provides an example, where `Pulitzer prizes' is an important entity and has occurred multiple times in multiple discourse units. The constructed Coreference Graph is shown on the right side of the document\footnote{We intentionally ignore other entities and mentions in this example for simplicity.}. 
When graph $\mathcal{G}_{C}$ is constructed, edges among 1-1, 2-1, 20-1 and 22-1 are all connected due to the mentions of `Pulitzer prizes'.

\section{\textsc{DiscoBert} Model}
% In this section, we present \textsc{DiscoBert}, a BERT-based extractive summarization model, which takes EDUs as the minimal selection unit for redundancy reduction and uses discourse graphs to capture long-range dependencies between EDUs. 
%with leverage of discourse units and discourse graphs.
% Although it is essentially a compressive model, the model is an end-to-end content selection where the minimal selection unit is an Elementary Discourse Unit (EDU). 
%We first provide an overview of the proposed method, then introduce the document encoder and graph encoder in detail.

\subsection{Overview}
Figure~\ref{fg:model} provides an overview of the proposed model, consisting of a Document Encoder and a Graph Encoder. For the
Document Encoder, a pre-trained BERT model is first used to encode the whole document on the token level. Then, a self-attentive span extractor is used to obtain the EDU representations from the corresponding text spans. The
Graph Encoder takes the output of the Document Encoder as input and updates the EDU representations with Graph Convolutional Network based on the constructed discourse graphs, which are then used to predict the oracle labels. 
%After the Graph Encoder, we score all of the discourse units based on the representations and make selection.
%\jcxu{PUT this into the caption of model?}

% The proposed model first takes prepossessed input document where the sentences are segmented into discourse units. 
% The dependencies between discourse units will be used in the decoding stage.
% Then we use Bert as the text encoder to encode the whole document. 
% After that, a self-attentive span extraction module is used to transform the sequence of token representations into a single discourse representation. 
% Given the representations of the discourse units and the constructed sparse discourse graphs, we use Graph Convolution Networks to encode and update the discourse representations.
% Finally, for every discourse unit, we predict a score to indicate its salience. 

Assume that document $D$ 
%consists of several sentences, and each sentence 
is segmented into $n$ EDUs in total, i.e.,
%\footnote{We will focus on EDU as the minimal selection unit (rather than sentence) in the following description because we only care about the sentence boundary when we pad a sentence with $\langle \text{CLS} \rangle$  and $\langle \text{SEP} \rangle$ in BERT.}
% We denote the $i$-th sentence and the $j$-th discourse with $s_i$ and $d_j$. 
% The whole document is represented as $D = \{ s_1, s_2, \cdots, s_m \} = \{ d_1, d_2, \cdots, d_n \}$. 
%The whole document is then represented as 
$D = \{ d_1, d_2, \cdots, d_n \}$, where $d_i$ denotes the $i$-th EDU.
%$d_i$ is represented as $d_i = \{ w_{i1}, w_{i2}, \cdots, w_{im} \}$, where $w_{ij}$ denotes the $j$-th token in the $i$-th EDU, and $m$ is the length of the EDU.
% The input of the \dbert model includes: 1) the whole document and tokenized by Bert tokenizer,
%
Following \citet{liu-lapata-2019-text}, we formulate extractive summarization as a sequential labeling task, where each EDU $d_i$ is scored by neural networks, and decisions are made based on the scores of all EDUs. 
The oracle labels are a sequence of binary labels, where 1 stands for being selected and 0 for not. 
We denote the labels as $Y=\{y_1^{*}, y_2^{*}, \cdots, y_n^{*} \}$.
During training, we aim to predict the sequence of labels $Y$ given the document $D$. 
During inference, we need to further consider discourse dependency to ensure the coherence and grammaticality of the output summary. 

%\vspace{-2mm}
\subsection{Document Encoder}
BERT is a pre-trained deep bidirectional Transformer encoder \cite{vaswani2017attention,devlin-etal-2019-bert}.
Following \citet{liu-lapata-2019-text}, we encode the whole document with BERT and finetune the BERT model for summarization. 

BERT is originally trained to encode a single sentence or sentence pair. However, a news article typically contains more than 500 words, hence we need to make some adaptation to apply BERT for document encoding. Specifically,
%\begin{enumerate}
we insert $\langle \text{CLS} \rangle$ and $\langle \text{SEP} \rangle$ tokens at the beginning and the end of each sentence, respectively.\footnote{We also tried inserting $\langle \text{CLS} \rangle$ and $\langle \text{SEP} \rangle$ at the beginning and the end of every EDU, and treating the  corresponding $\langle \text{CLS} \rangle$ representation as the representation for each EDU, but the performance drops drastically.} 
%The possible reason is that each sentence is segmented into around 4 EDUs by average so it is very different from the original BERT setting.
%
In order to encode long documents such as news articles, we also extend the maximum sequence length that BERT can take from 512 to 768 in all our experiments.
%In the original BERT model, the maximum length is 512 and the length of the pre-trained position embedding is also 512, which is insufficient for  
%Preliminary experiments show that neither linear interpolation of existing position embedding nor pasting the existing position embedding to the extended part work. Initialization with zero or with Gaussian distribution actually works. In practice we extend the length to 768 BPE for all the experiments.
%\end{enumerate}

The input document after tokenization is denoted as $D = %\{ w_{1S}, w_{11}, \cdots, w_{il_{i}}, w_{iE}, w_{(i+1)S}, w_{(i+1)1},  \cdots, w_{nE} \}$, 
\{d_1,\cdots,d_n \}$, and $d_i = \{w_{i1}, \cdots, w_{i\ell_{i}}\}$, 
where $\ell_{i}$ is the number of BPE tokens in the $i$-th EDU. If $d_i$ is the first EDU in a sentence, there is also a $\langle \text{CLS} \rangle$ token prepended to $d_i$; if $d_j$ is the last EDU in a sentence, there is a $\langle \text{SEP} \rangle$ token appended to $d_j$ (see Figure~\ref{fg:model}). The schema of insertion of $\langle \text{CLS} \rangle$ and $\langle \text{SEP} \rangle$ is an approach used in \citet{liu-lapata-2019-text}. 
 For simplicity, these two tokens are not shown in the equations. BERT model is then used to encode the document: 
%The S and E here stand for S(tart) and E(nd) respectively. 
%Formally we write it as:
% $ \{ h_{1S}, \cdots, h_{nE}\} = \text{BERT}(\{ w_{1S}, \cdots, w_{nE}\}) $
%
\begin{align*}
 \{ \mathbf{h}_{11}^{B}, \cdots, \mathbf{h}^{B}_{n\ell_n}  \} = \text{BERT}(\{ w_{11}, \cdots, w_{n\ell_n}\})\,,
    % h_{1S}, h_{11}, \cdots, h_{il_{i}}, h_{iE}, w_{(i+1)S}, w_{(i+1)1},  \cdots, w_{nE} \\
    % = \text{BERT}(w_{1S}, w_{11}, \cdots, w_{il_{i}}, w_{iE}, w_{(i+1)S}, w_{(i+1)1},  \cdots, w_{nE})\\
%   \text{score}_{t,i} = W_m \text{tanh} (W_d d_t + W_h h_i)\\
   %\hat{s}_t = \text{argmax}(\text{score}_{t,i})\\
%   p(\hat{s}_t=s_i|d_t,h_{k},v_{doc},h_i)= \text{softmax}(\text{score}_{t,i})
\end{align*}
where $\{ \mathbf{h}_{11}^{B}, \cdots, \mathbf{h}^{B}_{n\ell_n} \}$ is the BERT output of the whole document in the same length as the input.

After the BERT encoder, the representation of the $\langle \text{CLS} \rangle$ token can be used as sentence representation. However, this approach does not work in our setting, since we need to extract the representation for EDUs instead.
Therefore, we adopt a Self-Attentive Span Extractor (SpanExt), proposed in \citet{lee-etal-2017-end}, to learn EDU representation. % in a more flexible fashion.

For the $i$-th EDU with $\ell_i$ words, with the output from the BERT encoder $\{\mathbf{h}^{B}_{i1}, \mathbf{h}^{B}_{i2}, \cdots, \mathbf{h}^{B}_{i{\ell_i}}\}$, we obtain EDU representation as follows:
\begin{align*}
  \small
    \mathbf{\alpha}_{ij} &= \mathbf{W}_{2} \cdot \text{ReLU}(\mathbf{W}_{1} \mathbf{h}^{B}_{ij} +\mathbf{b}_{1}) +\mathbf{b}_{2}  \\
     \mathbf{a}_{ij} &= \frac{  \exp ({\mathbf{ \alpha}_{ij} }) }  {\sum_{k=1}^{\ell_i} \exp ({ \mathbf{\alpha}_{ik}  })        }\,, \quad 
     \mathbf{h}_{i}^{S} = \sum_{j=1}^{\ell_i} \mathbf{a}_{ij}  \cdot \mathbf{h}^{B}_{ij}  \,, 
     \end{align*}
where $\mathbf{\alpha}_{ij}$ is the score of the $j$-th word in the EDU, $\mathbf{a}_{ij}$ is the normalized attention of the $j$-th word w.r.t. all the words in the span. $\mathbf{h}_{i}^{S}$ is a weighted sum of the BERT output hidden states. Throughout the paper, all the $\mathbf{W}$ matrices and $\mathbf{b}$ vectors are parameters to learn. We abstract the above Self-Attentive Span Extractor as $\mathbf{h}_{i}^{S} = \text{SpanExt}( \mathbf{h}_{i1}^{B}, \cdots, \mathbf{h}_{i\ell_i}^{B} )$.
% \begin{align}
%   \mathbf{h}_{i}^{\text{S}} = \text{SpanExt}( \mathbf{h}_{i1}^{\text{B}}, \cdots, \mathbf{h}_{i\ell_i}^{\text{B}} ) \,.
% \end{align}

After the span extraction step, the whole document is represented as a sequence of EDU representations: $\mathbf{h}^{S} = \{ \mathbf{h}_1^{S}, \cdots, \mathbf{h}_n^{S} \} \in \mathbb{R}^{d_h \times n}$, which will be sent to the graph encoder.

\subsection{Graph Encoder}
Given the constructed graph $\mathcal{G}=(\mathcal{V}, \mathcal{E})$, nodes $\mathcal{V}$ correspond to the EDUs in a document, and edges $\mathcal{E}$ correspond to either RST discourse relations or coreference mentions. We then use Graph Convolutional Network to update the representations of all the EDUs, to capture long-range dependencies missed by BERT for better summarization. 
To modularize architecture design, we present a single Discourse Graph Encoder (DGE) layer. Multiple DGE layers are stacked in our experiments.

Assume that the input for the $k$-th DGE layer is denoted as $\mathbf{h}^{(k)} = \{\mathbf{h}^{(k)}_1,\ldots,\mathbf{h}^{(k)}_n\}\in \mathbb{R}^{d_h \times n}$, and the corresponding output is denoted as $\mathbf{h}^{(k+1)} = \{\mathbf{h}^{(k+1)}_1,\ldots,\mathbf{h}^{(k+1)}_n\}\in \mathbb{R}^{d_h \times n}$. 
The $k$-th DGE layer is designed as follows:
\begin{align*}
    \small
    \mathbf{u}_{i}^{(k)} &= \mathbf{W}_{4}^{(k)} \text{ReLU}(\mathbf{W}_{3}^{(k)}  \mathbf{h}_{i}^{(k)} +\mathbf{b}_{3}^{(k)}) +\mathbf{b}_{4}^{(k)}  \\ 
    \mathbf{v}_{i}^{(k)} &= \text{LN}( \mathbf{h}_{i}^{(k)} + \text{Dropout}(\mathbf{u}_{i}^{(k)})) \\
    \mathbf{w}_i^{(k)} &= \text{ReLU}\Big(\sum_{j \in \mathcal{N}_i} \frac{1}{|\mathcal{N}_i|} \mathbf{W}_{5}^{(k)} \mathbf{v}_j^{(k)} +\mathbf{b}_{5}^{(k)} \Big)\\
    \mathbf{h}_{i}^{(k+1)} &= \text{LN}(\text{Dropout}(\mathbf{w}_i^{(k)}) + \mathbf{v}_{i}^{(k)} )\,,
\end{align*}
where LN$(\cdot)$ represents Layer Normalization, $\mathcal{N}_i$ denotes the neighorhood of the $i$-th EDU node. $\mathbf{h}_{i}^{(k+1)}$ is the output of the $i$-th EDU in the $k$-th DGE layer, and $\mathbf{h}^{(1)} = \mathbf{h}^{S}$, which is the output from the Document Encoder. After $K$ layers of graph propagation, we obtain $\mathbf{h}^{G} = \mathbf{h}^{(K+1)} \in \mathbb{R}^{d_h \times n}$, which is the final representation of all the EDUs after the stacked DGE layers.
For different graphs, the parameter of DGEs are not shared. 
If we use both graphs, their output are concatenated: $ \mathbf{h}^{G} = \text{ReLU}(\mathbf{W}_{6} [\mathbf{h}_{C}^{G}; \mathbf{h}_{R}^{G}] + \mathbf{b}_{6})\,.$
% \begin{align*}
% \small
%     \mathbf{h}^{G} = \text{ReLU}(\mathbf{W}_{6} [\mathbf{h}_{C}^{G}; \mathbf{h}_{R}^{G}] + \mathbf{b}_{6})\,.
% \end{align*}

%\jcxu{talk about the detail about dge: mean }

%\jcxu{fusion of two graphs}

\subsection{Training \& Inference}
During training, $\mathbf{h}^{G}$ is used for predicting the oracle labels.
% If we use both two graphs, $\mathbf{h}^{G}$ is the concatenation of the two Graph Encoder outputs. 
Specifically, $ \hat{y}_i = \sigma (\mathbf{W}_7 \mathbf{h}_{i}^{G} +\mathbf{b}_7)$
%
%For the \textsc{DiscoBert} model without graph layers, we compute the scores of all EDUs given $H_{\text{SPAN}}$ with an MLP layer:
where $\sigma(\cdot)$ represents the logistic function, and $\hat{y}_i$ is the prediction probability ranging from 0 to 1. 
The training loss of the model is binary cross-entropy loss given the predictions and oracles: $  \mathcal{L}  = -\sum_{i=1}^{n}  ( y_{i}^{*} \log ( \hat{y}_i  ) + (1- y_{i}^{*} ) \log (1 - \hat{y}_i ) )\,.$
% \begin{align*}
%   \mathcal{L}  = -\sum_{i=1}^{n}  ( y_{i}^{*} \log ( \hat{y}_i  ) + (1- y_{i}^{*} ) \log (1 - \hat{y}_i ) )\,.
% \end{align*}
% The loss above is summed over all the training samples. 
For \textsc{DiscoBert} without graphs, the output from Document Encoder $\mathbf{h}^{S}$ is used for prediction instead. The creation of oracle is operated on EDU level. We greedily pick up EDUs with their necessary dependencies until R-1 F$_{1}$ drops.

During inference, given an input document, after obtaining the prediction probabilities of all the EDUs, i.e., $\hat{\mathbf{y}} = \{ \hat{y}_1, \cdots, \hat{y}_n\}$, 
we sort $\hat{\mathbf{y}}$ in descending order, and select EDUs accordingly. Note that the dependencies between EDUs are also enforced in prediction to ensure grammacality of generated summaries. 

%\jcxu{say some thing about oracle}

% The dependencies between EDUs are reflected in the construction of oracle summaries during training, and enforced during inference to ensure grammaticality.\jj{This seems redundant to the last sentence in the previous paragraph?}
\section{Experiments}
In this section, we present experimental results on two popular news summarization datasets. We compare our proposed model with state-of-the-art baselines and conduct detailed analysis to validate the effectiveness of \textsc{DiscoBert}.
%introduce the experimental setup, competitor models, implementation details, and present results on CNNDM and NYT.

% \begin{table}
% \centering
% \small
% \begin{tabular}{l|cc}
% \toprule
%                   & CNNDM  & NYT     \\
% \midrule
% \# sentences in Doc          & 24     & 22      \\
% \# EDUs in Doc          & 67     & 66      \\
% \# tokens in Doc          & 541    & 591     \\
% \# tokens in Sum     & 54     & 87      \\
% \midrule

% \# Train             & 287,226 & 137,778  \\
% \# Dev               & 13,368  & 17,222   \\
% \# Test              & 11,490  & 17,223  \\
% \midrule
% \# $\mathcal{E}$ in $\mathcal{G}_{D}$   & 66     & 65      \\
% \# $\mathcal{E}$ in $\mathcal{G}_{C}$  & 233    & 143     \\
% \bottomrule
% \end{tabular}
% \caption{Statistics of the dataset. In the first section we show the average number of sentences, EDUs and tokens in the document, and tokens in the reference summary. 
% The second section shows the dataset split.
% The third section shows the average number of edges in RST Discourse Graphs and Coreference Mention Graphs.
% % In CM Graphs all of the graph edges are bidirectional 
% }
% \end{table}

\begin{table}[t!]
\centering
\setlength{\tabcolsep}{4pt}
\small
\begin{tabular}{c|ccc|c|cc}
\toprule
    \multicolumn{1}{c|}{\multirow{2}{*}{Dataset}} & \multicolumn{3}{c|}{Document}                                                 & \multicolumn{1}{c|}{Sum.} & \multicolumn{2}{c}{\#~$\mathcal{E}$ in Graph}                                    \\
\multicolumn{1}{c|}{}                         & \multicolumn{1}{c}{\# sent.} & \multicolumn{1}{c}{\# EDU} & \multicolumn{1}{c|}{\# tok.} & \multicolumn{1}{c|}{\# tok.} & \multicolumn{1}{c}{  $\mathcal{G}_{R}$  } & \multicolumn{1}{c}{$\mathcal{G}_{C}$} \\\midrule
CNNDM &   24      & 67      & 541      &    54 &    66            &    233      \\
NYT   &    22     &     66  &   591    &    87 &        65        &     143    \\\bottomrule
\end{tabular}
\caption{Statistics of the datasets. The first block shows the average number of sentences, EDUs and tokens in the documents. 
The second block shows the average number of tokens in the reference summaries.
The third block shows the average number of edges in the constructed RST Graphs ($\mathcal{G}_{R}$) and Coreference Graphs ($\mathcal{G}_{C}$), respectively.}
% In CM Graphs all of the graph edges are bidirectional 
\label{tb:data}
%\vspace{-3mm}
\end{table}
\subsection{Datasets}
We evaluate the models on two datasets: New York Times (NYT)~\cite{Sandhaus_The_2008}, CNN and Dailymail (CNNDM)~\cite{Hermann_Teaching_2015}. 
We use the script from \citet{See_Get_2017} to extract summaries from raw data, and Stanford CoreNLP for sentence boundary detection, tokenization and parsing \cite{Manning_The_2014}. 
Due to the limitation of BERT, we only encode up to 768 BERT BPEs.

Table~\ref{tb:data} provides statistics of the datasets. The edges in  $\mathcal{G}_{C}$ are undirected, while those in $\mathcal{G}_{R}$ are directional.  
For CNNDM, there are 287,226, 13,368 and 11,490 samples for training, validation and test, respectively. We use the un-anonymized version as in previous summarization work.
NYT is licensed by LDC\footnote{\url{https://catalog.ldc.upenn.edu/LDC2008T19} }.
% Following \cite{Durrett_Learning_2016}, we discard the documents with the summary shorter than 50 words. 
Following previous work \cite{zhang-etal-2019-hibert,xu-durrett-compression}, we use 137,778, 17,222 and 17,223 samples for training, validation, and test, respectively. 
%The split is 

% \# Train             & 287,226 & 137,778  \\
% \# Dev               & 13,368  & 17,222   \\
% \# Test              & 11,490  & 17,223  \\
% \subsection{Experimental Setup}

\subsection{State-of-the-art Baselines}
We compare our model with the following state-of-the-art neural text summarization models. 

\vspace{1.5mm}
\noindent \textbf{Extractive Models:}
\textbf{BanditSum} treats extractive summarization as a contextual bandit problem, trained with policy gradient methods \cite{Dong_BanditSum_2018}.
\textbf{NeuSum} is an  extractive model with seq2seq architecture, where the attention mechanism scores the document and emits the index as the selection \cite{Zhou_Neural_2018}.
% \textbf{DeepChannel} is an extractive model with salience estimation and contrastive training strategy \cite{shi2019deepchannel}.

\vspace{1.5mm}
\noindent \textbf{Compressive Models:}
\textbf{JECS} is a neural text-compression-based summarization model using BLSTM as the encoder \cite{xu-durrett-compression}. The first stage is selecting sentences, and the second stage is sentence compression by pruning constituency parsing tree.

\vspace{1.5mm}
\noindent \textbf{BERT-based Models:} BERT-based models have achieved significant improvement on CNNDM and NYT, when compared with LSTM counterparts. \textbf{BertSum} is the first BERT-based extractive summarization model \cite{liu-lapata-2019-text}. Our baseline model \textsc{Bert} is the re-implementation of BertSum.
\textbf{PNBert} proposed a BERT-based model with various training strategies, including reinforcement learning and Pointer Networks \cite{zhong-etal-2019-searching}.
\textbf{HiBert} is a hierarchical BERT-based model for document encoding, which is further pretrained with unlabeled data \cite{zhang-etal-2019-hibert}.

\begin{table}[t!]
\centering
\small
\setlength{\tabcolsep}{4pt}
\begin{tabular}{l|ccc}
\toprule
Model                                 & R-1   & R-2   & R-L    \\
\midrule
Lead3  & 40.42 & 17.62 &  36.67\\
Oracle (Sentence) & 55.61 & 32.84 & 51.88 \\
Oracle (Discourse) &61.61 & 37.82 & 59.27  \\
\midrule
NeuSum \cite{Zhou_Neural_2018} & 41.59 & 19.01 &37.98 \\
BanditSum \cite{Dong_BanditSum_2018} & 41.50 &18.70& 37.60\\
JECS \cite{xu-durrett-compression} & 41.70 & 18.50& 37.90 \\

PNBERT   \cite{zhong-etal-2019-searching}       & 42.39 & 19.51 & 38.69  \\
PNBERT w. RL                            & 42.69 & 19.60 & 38.85  \\
BERT \cite{zhang-etal-2019-hibert}  &41.82  &19.48 &38.30 \\
$\text{HIBERT}_{S}$  & 42.10 & 19.70 & 38.53  \\
$\text{HIBERT}_{S}^{*}$   & 42.31 & 19.87 & 38.78  \\  
%  (w. more pretraining data) 
$\text{HIBERT}_{M}^{*}$  & 42.37 & 19.95 & 38.83  \\
% (w. more pretraining data)
BERTSUM  \cite{liu-lapata-2019-text}    & \textbf{43.25} & 20.24 & \textbf{39.63}  \\
% UNILM \cite{dong2019unified} & \textbf{43.33} & \textbf{20.21} & \textbf{40.51}\\
T5-Base \cite{raffel2019exploring} & 42.05 & \textbf{20.34} & 39.40 \\  % results on page 31 of https://arxiv.org/pdf/1910.10683.pdf
\midrule
\textsc{Bert}   & 43.07 & 19.94 & 39.44  \\
\textsc{DiscoBert}                             & 43.38 & 20.44 & 40.21  \\
\textsc{DiscoBert}  w. $\mathcal{G}_{C}$                      & 43.58 & 20.64 & 40.42  \\
\textsc{DiscoBert}  w. $\mathcal{G}_{R}$                     & 43.68 & 20.71 & 40.54  \\
\textsc{DiscoBert}  w.   $\mathcal{G}_{R}$ \&  $\mathcal{G}_{C}$                         & \textbf{43.77} & \textbf{20.85} & \textbf{40.67}  \\
\bottomrule
\end{tabular}
\vspace{-2mm}
\caption{Results on the test set of the CNNDM dataset. ROUGE-1, -2 and -L $\text{F}_{1}$ are reported. Models with the asterisk symbol (*) used extra data for pre-training. R-1 and R-2 are shorthands for unigram and bigram overlap; R-L is the longest common subsequence.}
\label{tb:cnndm}
%\vspace{-3mm}
\end{table}

\subsection{Implementation Details}
\label{sec:impl}
We use AllenNLP \cite{gardner-etal-2018-allennlp} as the code framework. The implementation of graph encoding is based on DGL \cite{wang2019dgl}.
Experiments are conducted on a single NVIDIA P100 card, and the mini-batch size is set to 6 due to GPU memory capacity. The length of each document is truncated to 768 BPEs.
We use the pre-trained `bert-base-uncased' model and fine tune it for all experiments. We train all our models for up to 80,000 steps. ROUGE~\cite{Lin2004} is used as the evaluation metrics, and `R-2' is used as the validation criteria.

%\paragraph{Preprocessing of EDUs}
% The preprocessing is mainly divided into two parts: segmentation and parsing of discourse units, and the construction of discourse graphs.

% \paragraph{Segmentation \& Parsing}
The realization of discourse units and structure is a critical part of EDU pre-processing, which requires two steps: discourse segmentation and RST parsing.
In the segmentation phase, we use a neural discourse segmenter based on the BiLSTM CRF framework \cite{wang-etal-2018-toward}\footnote{\url{https://github.com/PKU-TANGENT/NeuralEDUSeg}}. The segmenter achieved 94.3 F$_{1}$ score on the RST-DT test set, in which the human performance is 98.3.
In the parsing phase, we use a shift-reduce discourse parser to extract relations and identify neuclrity \cite{ji-eisenstein-2014-representation}\footnote{\url{https://github.com/jiyfeng/DPLP}}. 
% Both models are trained on Wall Street Journal articles from Penn Treebank. 

The dependencies among EDUs are crucial to the grammaticality of selected EDUs.
Here are the two steps to learn the derivation of dependencies: \textit{head inheritance} and \textit{tree conversion}. 
Head inheritance defines the head node for each valid non-terminal tree node. For each leaf node, the head is itself. 
We determine the head node(s) of non-terminal nodes based on their nuclearity.\footnote{If both children are N(ucleus), then the head of the current node inherits the head of the left child. Otherwise, when one child is N and the other is S, the head of the current node inherits the head of the N child.}
For example, in Figure~\ref{fig:disco}, the heads of text spans [1-5], [2-5], [3-5] and [4-5] need to be grounded to a single EDU.
% In RST, each EDU is tagged as either \textit{Nucleus} or \textit{Satellite}, which characterizes the document's nuclearity or saliency. \jj{Should explain what is nuclearity or saliency} \textit{Nucleus} (N) nodes are more central in terms of its content and grammatical/lexical reliance, while \textit{Satellite} (S) nodes are more peripheral. 
We propose a simple yet effective schema to convert RST discourse tree to a dependency-based discourse tree.\footnote{If one child node is N and the other is S, the head of the S node depends on the head of the N node. 
If both children are N and the right child does not contain a subject in the discourse, the head of the right N node depends on the head of the left N node.}
% , which covers most cases without prohibiting reasonable operations, compared to the approaches proposed by 
% \citet{hirao-etal-2013-single,Durrett_Learning_2016}. Detailed comparison of different conversion methods can be found in the Appendix.
% The way we convert the trees is summarized as follows:
% \jj{What's the relation of this paragraph and the one earlier on N and S nodes inheritance?}
% After the derivation of dependency-based discourse tree, we obtain the dependency relations among all EDUs. 
We always consider the dependency restriction such as the reliance of Satellite on Nucleus, when we create oracle during pre-processing and when the model makes the prediction. 
For the example in Figure~\ref{fig:disco}, if the model selects \say {[5] being carried ... Liberia.} as a candidate span, we will enforce the model to select \say{[3] and shows ... 8,} and \say{[2] This ... series,} as well.
% (the dependencies in this example are \{($4\rightarrow5$), ($5\rightarrow3$), ($3\rightarrow2$), ($1\rightarrow2$)\}).

% since we have a clear schema from discourse segmentation to dependency resolution to guarantee the grammaticality and factuality. 
The number of chosen EDUs depends on the average length of the reference summaries, dependencies across EDUs as mentioned above, and the length of the existing content. The optimal average number of EDUs selected is tuned on the development set. 

\subsection{Experimental Results}

% \vspace{1.5mm}
% \noindent \textbf{Results on CNNDM}
\paragraph{Results on CNNDM}
Table~\ref{tb:cnndm} shows results on CNNDM. 
The first section includes Lead3 baseline, sentence-based oracle, and discourse-based oracle. 
The second section lists the performance of baseline models, including non-BERT-based and BERT-based variants.
The performance of our proposed model is listed in the third section. \textsc{Bert} is our implementation of sentence-based BERT model. 
\textsc{DiscoBert} is our discourse-based BERT model without Discourse Graph Encoder.
\textsc{DiscoBert} w. $\mathcal{G}_{C}$ and \textsc{DiscoBert} w. $\mathcal{G}_{R}$ are the discourse-based BERT model with Coreference Graph and RST Graph, respectively. 
\textsc{DiscoBert}  w.   $\mathcal{G}_{R}$ \&  $\mathcal{G}_{C}$ is the fusion model encoding both graphs.

\begin{table}[t]
\centering
\setlength{\tabcolsep}{4pt}
\small
\begin{tabular}{l|ccc}
\toprule
Model                            & R-1  & R-2  & R-L   \\
\midrule
Lead3                             & 41.80 & 22.60 & 35.00    \\
Oracle (Sentence)          & 64.22 & 44.57 & 57.27    \\
Oracle (Discourse)          & 67.76 & 48.05 & 62.40    \\
\midrule
% BLSTM Extraction                 & 44.3 & 25.5 & 37.1  \\
JECS    \cite{xu-durrett-compression}                         & 45.50 & 25.30 & 38.20  \\
BERT        \cite{zhang-etal-2019-hibert}                     & 48.38 & 29.04   & 40.53  \\
$\text{HIBERT}_{S}$                 & 48.92 & 29.58 & 41.10  \\
$\text{HIBERT}_{M}$                        & 49.06 & 29.70 & 41.23  \\
$\text{HIBERT}_{S}^{*}$   & 49.25 & 29.92 & 41.43  \\
$\text{HIBERT}_{M}^{*}$   & \textbf{49.47} &\textbf{ 30.11} & \textbf{41.63 } \\
\midrule
\textsc{Bert}                             & 48.48 & 29.01   & 40.62  \\
\textsc{DiscoBert}                        & 49.78 & 30.30 & 42.44  \\
\textsc{DiscoBert}  w. $\mathcal{G}_{C}$                   & 49.79 & 30.18 & 42.48  \\
\textsc{DiscoBert}  w. $\mathcal{G}_{R}$                    & 49.86   & 30.25 & 42.55  \\
\textsc{DiscoBert}   w. $\mathcal{G}_{R}$ \& $\mathcal{G}_{C}$   & \textbf{50.00}   & \textbf{30.38}    & \textbf{42.70}    \\\bottomrule
\end{tabular}
\caption{Results on the test set of the NYT dataset. Models with the asterisk symbol (*) used extra data for pre-training.}
\label{tb:nyt}
%\vspace{-3mm}
\end{table}

The proposed \textsc{DiscoBert} beats the sentence-based counterpart and all the competitor models. With the help of Discourse Graph Encoder, the graph-based \textsc{DiscoBert} beats the state-of-the-art BERT model by a significant margin (0.52/0.61/1.04 on R-1/-2/-L on F$_{1}$). Ablation study with individual graphs shows that the RST Graph is slightly more helpful than the Coreference Graph, while the combination of both achieves better performance overall. 

% \vspace{1.5mm}
% \noindent \textbf{Results on NYT}
\paragraph{Results on NYT}
Results are summarized in Table~\ref{tb:nyt}. The proposed model surpasses previous state-of-the-art BERT-based model by a significant margin. HIBERT$_{S}^{*}$ and HIBERT$_{M}^{*}$ used extra data for pre-training the model.
We notice that in the NYT dataset, most of the improvement comes from the use of EDUs as minimal selection units. \textsc{DiscoBert} provides 1.30/1.29/1.82 gain on R-1/-2/-L over the \textsc{Bert} baseline. However, the use of discourse graphs does not help much in this case. 

%\section{Discussion}
\subsection{Grammaticality}
% \vspace{1.5mm}
% \noindent \textbf{Grammaticality}
Due to segmentation and partial selection of sentence, the output of our model might not be as grammatical as the original sentence. 
We manually examined and automatically evaluated model output, and observed that overall, the generated summaries are still grammatical, given the RST dependency tree constraining the rhetorical relations among EDUs. A set of simple yet effective post-processing rules helps to complete the EDUs in some cases.
\paragraph{Automatic Grammar Checking}
We followed \citet{xu-durrett-compression} to perform automatic grammar checking using Grammarly. 
Table~\ref{tb:grm} shows the grammar checking results, where the average number of errors in every 10,000 characters on CNNDM and NYT datasets is reported. We compare \textsc{DiscoBert} with sentence-based \textsc{Bert} model. `All' shows the summation of the number of errors in all categories. As shown in the table, the summaries generated by our model have retained the quality of the original text. 
% Our model performs worse in terms of punctuation, but improves the readability significantly, because the average length of each output sentence decreases. 

% \usepackage{multirow}

\begin{table}
\centering
\small
\begin{tabular}{cc|ccccc}
\toprule
Source                 & M & All  & CR & PV & PT & O  \\\midrule
\multirow{2}{*}{CNNDM} & Sent  & 33.0 & 18.7        & 9.0                  & 2.3                                                     & 3.0     \\
                       & Disco & 34.0 & 18.3        & 8.4                  & 2.6                                                   & 4.7     \\\midrule
\multirow{2}{*}{NYT}   & Sent  &  23.3   & 13.5& 5.9                      &                                           0.8                 &        3.1 \\
                       & Disco &   23.8   &         13.9    & 5.7                     & 0.8                                                       &       3.4
                      \\ \bottomrule
\end{tabular}
\caption{Number of errors per 10,000 characters based on automatic grammaticality checking with Grammarly on CNNDM and NYT. Lower values are better.
Detailed error categories, including correctness (CR), passive voice (PV) misuse, punctuation (PT) in compound/complex sentences and others (O), are listed from left to right.
% CoRrectness   Passive Voice Misuse     Punctuation in Compound/Complex Sentences   Hard-to-read Text O(thers) 
}
\label{tb:grm}
%\vspace{-1mm}
\end{table}

\begin{table}
\centering
\small
\begin{tabular}{c|ccc}
\toprule
Model     & All & Coherence  & Grammaticality   \\\midrule
Sent &  $3.45 \pm 0.87$ & $3.30 \pm 0.90$ & $3.45 \pm 1.06$ \\
Disco & $3.24 \pm 0.84$ & $3.15 \pm 0.95$ & $3.25 \pm 1.02$ \\
Ref  & $3.28 \pm 0.99$ & $3.12 \pm 0.94$ & $3.29\pm 1.06$ \\ \bottomrule
\end{tabular}
\caption{Human evaluation results. We ask Turkers to grade the overall preference, coherence and grammaticality from 1 to 5. Mean values along with standard deviations are reported.
% CoRrectness   Passive Voice Misuse     Punctuation in Compound/Complex Sentences   Hard-to-read Text O(thers) 
}
\label{tb:human}
%\vspace{-3mm}
\end{table}

\paragraph{Human Evaluation}
We sampled 200 documents from the test set of CNNDM and for each sample, we asked two Turkers to grade three summaries from 1 to 5. Results are shown in Table~\ref{tb:human}. Sent-BERT model (the original BERTSum model) selects sentences from the document, hence providing the best overall readability, coherence, and grammaticality. In some cases, reference summaries are just long phrases, so the scores are slightly lower than those from the sentence model. \textsc{DiscoBERT} model is slightly worse than Sent-BERT model but is fully comparable to the other two variants. 

\paragraph{Examples \& Analysis}
We show some examples of model output in Table~\ref{tb:examples}. We notice that a decent amount of irrelevant details are removed from the extracted summary.

Despite the success, we further conducted error analysis and found that the errors mostly originated from the RST dependency resolution and the upstream parsing error of the discourse parser.
The misclassification of RST dependencies and the hand-crafted rules for dependency resolution hurted the grammaticality and coherence of the `generated' outputs.
Common punctuation issues include extra or missing commas, as well as missing quotation marks. 
% For example, if we only select the first EDU of the sentence \say{[`Johnny is believed to have drowned,]$_{1}$ [but actually he is fine,']$_{2}$ [the police say.]$_{3}$}, the output \say{`Johnny is believed to have drowned.} does not look like a grammatical sentence due to the punctuation. 
Some of the coherence issue originates from missing or improper or missing anaphora  resolution. 
In this example \say{[`Johnny is believed to have drowned,]$_{1}$ [but actually \textit{he} is fine,']$_{2}$ [the police say.]$_{3}$}, only selecting the second EDU yields a sentence \say{actually he is fine}, which is not clear who is `he' mentioned here.

\begin{table}[]
\centering
\small
% \begin{tabular}{ll}
\begin{tabular}{p{0.45\textwidth}}
%   |p{\dimexpr.5\linewidth-2\tabcolsep-1.3333\arrayrulewidth}% column 1
%   |p{\dimexpr.25\linewidth-2\tabcolsep-1.3333\arrayrulewidth}% column 2
% \begin{tabularx}{\textwidth}{l|l}
\toprule
 Clare Hines \sout{, who lives in Brisbane,} was diagnosed with a brain tumour after suffering epileptic seizures. \sout{After a number of tests doctors discovered} she had a benign tumour \sout{that had wrapped itself around her acoustic, facial and balance nerve – and told her she had have it surgically removed or she risked the tumour turning malignant.} One week before brain surgery she found out she was pregnant. 
% https://www.dailymail.co.uk/news/article-3020974/Clare-s-choice-Doctors-warned-epileptic-mother-pregnancy-cause-brain-tumour-grow-pictures-prove-right-decision.html
 \\\midrule
% Jordan Henderson is 'over the moon' after signing a new long-term contract with Liverpool. 
Jordan Henderson\sout{, in action against Aston Villa at Wembley on Sunday,} has agreed a new Liverpool deal.
The club's vice captain puts pen to paper on a deal which will keep him at Liverpool until 2020. 
Rodgers will consider Henderson for the role of club captain \sout{after Steven Gerrard moves to LA Galaxy at the end of the campaign but, for now, the England international is delighted to have agreed terms on a contract that will take him through the peak years of his career}.
  \\\bottomrule   \end{tabular}
\caption{Example outputs from CNNDM by \textsc{DiscoBert}. Strikethrough indicates discarded EDUs.}
\label{tb:examples}
% \end{tabular}
%\vspace{-4mm}
\end{table}

\section{Related Work}
\paragraph{Neural Extractive Summarization}
Neural networks have been widely used in extractive summarization.
Various decoding approaches, including ranking \cite{Narayan_Ranking_2018}, index prediction \cite{Zhou_Neural_2018} and sequential labelling \cite{Nallapati_SummaRuNNer_2017,Zhang_Neural_2018,Dong_BanditSum_2018}, have been applied to content selection.
% \cite{Narayan_Ranking_2018} treated the problem as ranking and trained the model from ROUGE reward.
%  \cite{Zhou_Neural_2018} used a seq2seq model to predict the index of selected sentences with pointer network.
% \cite{Nallapati_SummaRuNNer_2017,Zhang_Neural_2018,Dong_BanditSum_2018} conceptualized the problem as a sequence labelling problem where each sentence is scored by neural networks.
% \paragraph{Pretrained LMs for Summarization}
% Fine-tuning BERT for text summarization has brought decent improvement \cite{zhong-etal-2019-searching,liu2019fine}.
Our model uses a similar configuration to encode the document with BERT as \citet{liu-lapata-2019-text} did, but we use discourse graph structure and graph encoder to handle the long-range dependency issue. 

\paragraph{Neural Compressive Summarization}
Text summarization with compression and deletion has been explored in some recent work. 
\citet{xu-durrett-compression} presented a two-stage neural model for selection and compression based on constituency tree pruning.
% \cite{cheng-lapata-2016-neural}
% \cite{Gehrmann_Bottom_2018}
\citet{dong-etal-2019-editnts} presented a neural sentence compression model with discrete operations including deletion and addition. Different from these studies, as we use EDUs as minimal selection basis, sentence compression is achieved automatically in our model.

\paragraph{Discourse \& Summarization}
%We have shown the viability of using EDU as the minimal selection unit rather than sentence.
% since we have a clear schema from discourse segmentation to dependency resolution to guarantee the grammaticality and factuality. 
%In the rest of the paper, we assume the selection unit is EDU unless specified. 
%
% \paragraph{Discourse in Summarization}
The use of discourse theory for text summarization has been explored before. 
\citet{louis-etal-2010-discourse} examined the benefit of graph structure provided by discourse relations for text summarization. 
\citet{hirao-etal-2013-single,yoshida-etal-2014-dependency} formulated the summarization problem as the trimming of the document discourse tree.
\citet{Durrett_Learning_2016} presented a system of sentence extraction and compression with ILP methods using discourse structure. 
\citet{Li_The_2016} demonstrated that using EDUs as units of content selection leads to stronger summarization performance. Compared with them, our proposed method is the first neural end-to-end summarization model using EDUs as the selection basis.  

%\zhe{shall we add a sentence to describe the key difference between the usage of EDU in our work, when compared with others?}
%\jcxu{the simple answer here is: we are the first neural model using discourse units in sum model. previous approaches are very different and not end-to-end trained. }

\paragraph{Graph-based Summarization}
Graph approach has been explored in text summarization over decades. LexRank introduced a stochastic graph-based method for computing relative importance of
textual units \cite{erkan2004lexrank}. 
\citet{yasunaga-etal-2017-graph} employed a GCN on the relation graphs with sentence embeddings obtained from RNN. 
\citet{tan-etal-2017-abstractive} also proposed graph-based attention in abstractive summarization model. 
\citet{fernandes2018structured} developed a framework to reason long-distance relationships for text summarization.

\section{Conclusion}
In this paper, we present \textsc{DiscoBert}, which uses discourse unit as the minimal selection basis to reduce summarization redundancy and leverages two types of discourse graphs as inductive bias to
capture long-range dependencies among discourse units. We validate the proposed approach on
two popular summarization datasets, and observe consistent improvement
over baseline models. For future work, we will explore better graph encoding methods, and apply discourse graphs to
other tasks that require long document encoding.

\section*{Acknowledgement} 
Thanks to Junyi Jessy Li, Greg Durrett, Yen-Chun Chen, and to the other members of the Microsoft Dynamics 365 AI Research team for the proof-reading, feedback and suggestions. 

\bibliography{jcxu_full}
\bibliographystyle{acl_natbib}

\clearpage
\appendix

\section{Appendix}
% \label{sec:appendix}

\begin{figure}[h]
\centering
% \justify
% {Example: [Police]\textsubscript{1} [investigating the case]\textsubscript{2}  [learned]\textsubscript{3}  [where 36-year-old Vittorio Arrigoni was being held]\textsubscript{4}  [and went to the location,]\textsubscript{5}  [where they found the body,]\textsubscript{6}  [the statement said.]\textsubscript{7} }
\includegraphics[width=0.45\textwidth]{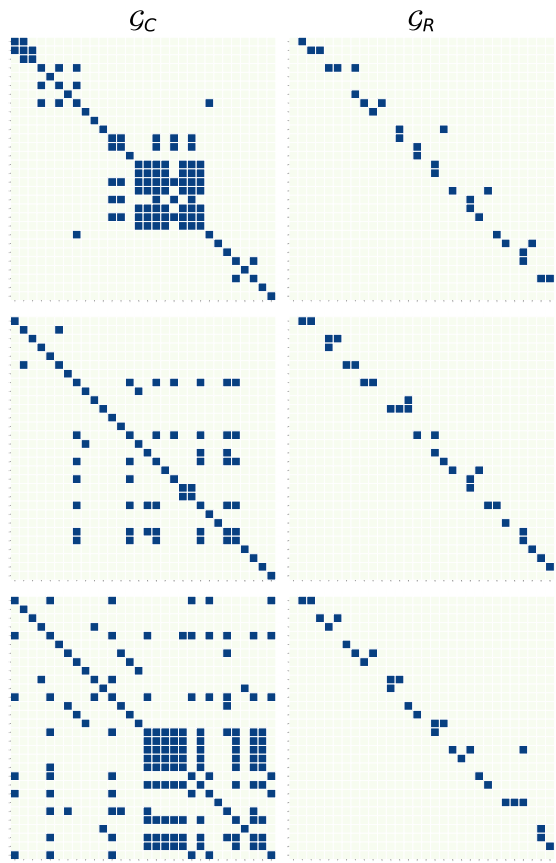}
\caption{Examples of the adjacent matrix of Coreference Graphs $\mathcal{G}_{C}$ and RST Graphs $\mathcal{G}_{R}$.
}
\label{fig:vismore}
\end{figure}

Figure~\ref{fig:vismore} provides three sets of examples of the constructed graphs from CNNDM. Specifically, $\mathcal{G}_{C}$ is strictly symmetric and self-loop is added to all the nodes to prevent the graph from growing too sparse.
On the other hand, all of the on-diagonal entries in $\mathcal{G}_{R}$ are zero because the node from RST graph never points to itself.

% \input{fig_graph_vis.tex}
% [width=0.48125\textwidth] for ACL
% [width=0.473\textwidth] for AAAI
% Police investigating the case learned where 36-year-old Vittorio Arrigoni was being held and went to the location, where they found the body, the statement said.

\end{document}